\documentclass[sigconf]{acmart}
\renewcommand\footnotetextcopyrightpermission[1]{} 
\usepackage{colortbl}
\usepackage{booktabs} 
\usepackage{amsmath}

\setcopyright{none}
\usepackage[normalem]{ulem} 

\usepackage{mathrsfs}
\usepackage{listings}
\usepackage{url}
\usepackage[algo2e,ruled,linesnumbered]{algorithm2e}
\usepackage{multirow}
\usepackage{array}
\usepackage{comment}

\usepackage{tikz}
\usetikzlibrary{trees}
\usetikzlibrary{calc}
\usetikzlibrary{arrows}
\usepackage{pgfplots}
\usetikzlibrary{shapes}
\usetikzlibrary{shapes.geometric}
\usetikzlibrary{arrows,decorations.pathmorphing,backgrounds,positioning,fit}
\usetikzlibrary{arrows,shapes,positioning,shadows,trees}
\usetikzlibrary{shapes,shadows,trees}

\begin{document}
\title{Local Rule-Based Explanations of Black Box Decision Systems}


\author{Riccardo Guidotti}
\affiliation{%
  \institution{ISTI-CNR \& University of Pisa, Italy}
}
\email{riccardo.guidotti@isti.cnr.it}

\author{Anna Monreale}
\affiliation{%
  \institution{University of Pisa, Italy}
}
\email{anna.monreale@unipi.it}

\author{Salvatore Ruggieri }
\affiliation{%
  \institution{University of Pisa, Italy}
}
\email{salvatore.ruggieri@unipi.it}

\author{Dino Pedreschi}
\affiliation{%
  \institution{University of Pisa, Italy}
}
\email{dino.pedreschi@unipi.it}

\author{Franco Turini}
\affiliation{%
  \institution{University of Pisa, Italy}
}
\email{franco.turini@unipi.it}

\author{Fosca Giannotti}
\affiliation{%
  \institution{ISTI-CNR of Pisa, Italy}
}
\email{fosca.giannotti@isti.cnr.it}

\renewcommand{\shortauthors}{R. Guidotti et. al.}

%

\begin{abstract}
The recent years have witnessed the rise of accurate but obscure decision systems which hide the logic of their internal decision processes to the users. 
The lack of explanations for the decisions of black box systems is a key ethical issue, and a limitation to the adoption of machine learning components in socially sensitive and safety-critical contexts.
In this paper we focus on the problem of black box outcome explanation, i.e., explaining the reasons of the decision taken on a specific instance.
We propose \textbf{LORE}, an agnostic method able to provide interpretable and faithful explanations.
\textbf{LORE} first leans a local interpretable predictor on a  synthetic neighborhood generated by a genetic algorithm.
Then it derives from the logic of the local interpretable predictor a meaningful explanation consisting of: a decision rule, which explains the reasons of the decision; and a set of counterfactual rules, suggesting the changes in the instance's features that lead to a different outcome. 
Wide experiments show that \textbf{LORE} outperforms existing methods and baselines both in the quality of explanations and in the accuracy in mimicking the black box.
\end{abstract}

%
%




\maketitle

\section{Introduction}
\label{sec:introduction}
Popular magazines and newspapers are full of commentaries about algorithms taking critical decisions that heavily impact on our life and society, from granting a loan to finding a job or driving our car. 
The worry is not only due to the increasing  automation of decision making, but mostly to the fact that the algorithms are opaque and their logic unexplained. 
The main cause for this lack of transparency is that often the algorithm itself has not been directly coded by a human but it has been generated from data through machine learning. 
Machine learning allows building predictive models which map user features into a class (outcome or decision), obtained by generalizing from a training set of examples.
This learning process is made possible by the digital records of past decisions and classification outcomes, typically provided by human experts and decision makers. 
The process of inferring a classification model from examples cannot be controlled step by step because the size of training data and the complexity of the learned model are too big for humans. 
This is how we got trapped in a paradoxical situation in which, on one side, the legislator defines new regulations requiring that automated decisions should be explained to affected people\footnote{We refer here to the so-called "right to explanation" established in the European General Data Protection Regulation (GDPR), entering into force in May 2018.} while, on the other side, even more sophisticated and obscure algorithms for decision making are generated \cite{wachter2017right,goodman2016eu}.

The lack of transparency in algorithms generated through machine learning grants to them the power to perpetuate or reinforce forms of injustice by learning bad habits from the data. 
In fact, if the training data contains a number of biased decision records, or misleading classification examples due to data collection mistakes or artifacts, it is likely that the resulting algorithm inherits the biases and recommends discriminatory or simply wrong decisions\footnote{\href{http://www.propublica.org/article/machine-bias-risk-assessments-in-criminal-sentencing}{www.propublica.org/article/machine-bias-risk-assessments-in-criminal-sentencing}} \cite{Barocas2016,Berk2017}. 
The inability of obtaining an explanation for what one considers a biased decision is a profound drawback of learning from big data, limiting social acceptance and trust on its adoption in many sensitive contexts. 
Starting from \cite{pedreshi2008discrimination} a rich literature has been flourishing on discrimination discovery and avoidance. 
Some of the ideas developed in that context can be reinterpreted for addressing the more general problem of explaining the logic driving a decision taken by an obscure algorithm, which is precisely the problem tackled in this paper.

In particular, in this paper we address the problem of explaining the decision outcome taken by an obscure algorithm by providing ``meaningful explanations of the logic involved'' when automated decision making takes place, as prescribed by the GDPR.
The decision system can be obscure because based on a deep learning approach, or because of inaccessibility of the source code, or other reasons.
We perform our research under some specific assumptions. 
First, we assume that an explanation is interesting for a user if it clarifies why a \textit{specific decision} pertaining that user has been made, i.e., we aim for \textit{local} explanations, not general, \textit{global}, descriptions of how the overall system works \cite{guidotti2018survey}. 
Second, we assume that the vehicle for offering explanations should be as close as possible to the language of reasoning, that is logic. 
Thus, we are also assuming that the user can understand elementary logic rules.
Finally, we assume that the black box decision system can be queried as many times as necessary, to probe its decision behavior to the scope of reconstructing its logic; this is certainly the case in a legal argumentation in court, or in an industrial setting where a company wants to stress-test a machine learning component of a manufactured product, to minimize risk of failures and consequent industrial liability. 
On the other hand, we make no assumptions on the specific algorithms used in the obscure classifier: we aim at an \textit{agnostic} explanation method, one that works analyzing the input-output behavior of the black box system, disregarding its internals. 

We propose a solution to the black box outcome explanation problem suitable for relational, tabular data, called \textbf{LORE} (for \textbf{LO}cal \textbf{R}ule-based \textbf{E}xplanations). 
Given a black box binary predictor $b$ and a specific instance $x$ labeled with outcome $y$ by $b$, we build a simple, interpretable predictor by first generating a balanced set of neighbor instances of the given instance $x$ through an ad-hoc genetic algorithm, and then extracting from such a set a decision tree classifier. 
A \textit{local explanation} is then extracted from the obtained decision tree. 
The local explanation is a pair composed by (\textit{i}) a \textit{logic rule}, corresponding to the path in the tree that explains why $x$ has been labeled as $y$ by $b$, and (\textit{ii}) a set of \textit{counterfactual rules}, explaining which conditions should be changed by $x$ so to invert the class $y$ assigned by $b$.
For example, from the \emph{compas} dataset \cite{Barocas2016,Berk2017} we may have the following explanation: the rule $\{\mathit{age}{\leq}39, \mathit{race}{=}\mathit{African{-}American}, \mathit{recidivist}{=}\mathit{True}\} {\rightarrow} \mathit{High\ Risk}$  and the counterfactuals $\{\mathit{age}{>}40\},\{ \mathit{race}{=}\mathit{Native{-}American}\}$. 

The intuition behind our method, common to other approaches, such as LIME \cite{ribeiro2016should}, and Anchor \cite{ribeiro2018anchors} is that the decision boundary for the black box can be arbitrarily complex over the whole data space, but in the neighborhood of a data point there is a high chance that the decision boundary is clear and simple, hence amenable to be captured by an interpretable model. 
The novelty of our method lies in (\textit{i}) a focused procedure, based on genetic algorithm, to explore the decision boundary in the neighborhood of the data point, which produces a high-quality training data to learn the local decision tree, and (\textit{ii}) a high expressiveness of the proposed local explanations, which surpasses state-of-the-art methods providing not only succinct evidence why a data point has been assigned a specific class, but also counterfactuals suggesting what should be different in the vicinity of the data point to reverse the predicted outcome. 
We propose extensive experiments to assess both quantitatively and qualitatively the accuracy of our explanation method.

In the rest of this paper, after describing the state of the art in the field of explanation of black box decision models (Section \ref{sec:related}), we offer a formalization of the problem by defining the notions of \textit{black box outcome explanation}, \textit{explanation through interpretable models}, and \textit{local explanation} (Section \ref{sec:problem}). 
We then define our method \textbf{LORE} in Section \ref{sec:method}. Section \ref{sec:experiments} is devoted to the experiments, the set up of which requires the definition of appropriate validation measures. 
We critically compare local versus global explanations, rule-based versus linear explanations, different types of rule-based explanations with respect to the state of the art, and discuss the advantages of genetic algorithms for neighborhood generation. 
Conclusions and future research directions are discussed in Section~\ref{sec:conclusion}.

\section{Related Work}
\label{sec:related}
Recently, the research of methods for explaining black box decision systems has caught much attention \cite{guidotti2018survey}.
A large number of papers propose approaches for understanding the \textit{global} logic of the black box by providing an 
interpretable classifier able to mimic the obscure decision system. 
Generally, these methods are designed for explaining specific black box models, i.e., they are not black box agnostic.
Decision trees have been adopted to globally explain neural networks \cite{craven1996extracting,krishnan1999extracting} and tree ensembles \cite{hara2016making,tan2016tree}.  
Classification rules have been widely used to explain neural networks \cite{johansson2004accuracy,augasta2012reverse,andrews1995survey} but also to understand the global behavior of SVMs \cite{fung2005rule,nunez2002rule}.
Only few methods for global explanation are agnostic with respect to the black box \cite{lou2012intelligible,henelius2014peek}.
In the cases in which the training set is available, classification rules are also widely used to avoid building black boxes by directly designing a transparent classifier \cite{guidotti2018survey} which is locally or \mbox{globally interpretable on its own \cite{wang2015falling,lakkaraju2016interpretable,malioutov2017learning}.}

Other approaches, more related to the one we propose, address the problem of explaining the \textit{local} behavior of a black box \cite{guidotti2018survey}.
In other words, they provide an explanation for the decision assigned to a specific instance. 
In this context there are two kinds of approaches: the model-dependent approaches and the agnostic ones. 
In the first category most of the papers aim at explaining neural networks and base their explanation on saliency masks, i.e., a subset of the instance that explains what is mainly responsible for the prediction \cite{xu2015show,zhou2016learning}.
Examples of salient mask are parts of an image, or words or sentences in a text.  
On the other hand, agnostic approaches provide explanations for any type of black box. 
In \cite{ribeiro2016should} the authors present LIME, which starts from instances randomly generated in the neighborhood of the instance to be explained. 
The method infers from them linear models as comprehensible local predictors. The importance of a feature in the linear model represents the explanation finally provided to the user.
As a limitation of the approach, a random generation of the neighborhood does not take into account density of black box outcomes in the neighborhood instances. Hence, the linear classifiers inferred from them may not correctly characterize outcome values as a function of the predictive features. We will instead use a genetic algorithm that exploits the black box for instance generation.

Extensions of LIME using decision rules (called anchors) and program expression trees are presented in \cite{ribeiro2018anchors} and \cite{singh2016programs} respectively. 
\cite{ribeiro2018anchors} uses a bandit algorithm that randomly constructs the anchors with the highest coverage and respecting a precision threshold.
\cite{singh2016programs} adopts a simulated annealing approach that randomly grows, shrinks, or replaces nodes in an expression tree.
%
%
The neighborhood generation process adopted is the same as in LIME. Another crucial weak point of those approaches, 
is the need for user-specified parameters for desired explanations:
the number of features \cite{ribeiro2016should}, the level of precision, the maximum expression tree depth \cite{ribeiro2018anchors}.  Our approach is instead parameter-free.




Concerning the counterfactual part of our notion of  explanation, \cite{wachter2017counterfactual} computes a counterfactual for an instance $x$ by solving an optimization problem over the space of instances. The solution is an instance $x'$ close to $x$ but with different outcome assigned by the black box\footnote{If instead of a black box, we are given a machine learning model, this problem is known as the \textit{inverse classification problem} \cite{DBLP:journals/jcst/AggarwalCH10}.}. Our approach provides a more abstract notion of counterfactuals, consisting of logic rules rather than flips of feature values. Thus, the user is given not only a specific example of how to obtain actionable recourse (e.g., how to improve application for getting a benefit), but also an abstract characterization of its neighboorhood instances with reversed black box outcome.

To the best of our knowledge, in the literature there is no work proposing a black box agnostic method for local decision explanation based on both decision and counterfactual rules.

\section{Problem and Explanations}
\label{sec:problem}
Let us start recalling basic notation on classification of tabular data. 
Afterwards, we define the \textit{black box outcome explanation problem}, and the notion of \textit{explanation} for which we propose a solution. 

\paragraph{Classification, black boxes, and interpretable predictors.} 
A \emph{predictor} or classifier, is a function $b:\mathcal{X}^{(m)} \rightarrow \mathcal{Y}$ which maps data instances (tuples) $x$ from a feature space  $\mathcal{X}^{(m)}$ with $m$ input features to a decision $y$ in a target space $\mathcal{Y}$. 
We write $b(x) = y$ to denote the decision $y$ predicted by $b$, and $b(X) = Y$ as a shorthand for $\{b(x) \ |\ x \in X\} = Y$. 
We restrict here to binary decisions.
An instance $x$ consists of a set of $m$ attribute-value pairs $(a_i, v_i)$, where $a_i$ is a feature (or attribute) and $v_i$ is a value from the domain of $a_i$. 
The domain of a feature can be continuous or categorical.
A predictor can be a machine learning model, a domain-expert rule-based system, or any combination of algorithmic and human knowledge processing. 
We assume that a predictor is available as a software function that can be queried at will.
In the following, we denote by $b$ a \emph{black box} predictor, whose internals are either unknown to the observer or they are known but uninterpretable by humans. 
Examples include neural networks, SVMs, ensemble classifiers, or a composition of data mining, legacy software, and hard-coded expert systems.
Instead, we denote with $c$ an \emph{interpretable} predictor, whose internal processing yielding a decision $c(x) = y$ can be given a symbolic interpretation understandable by a human. 
Examples of such predictors include rule-based classifiers, decision trees, decision sets, and rational functions.

\paragraph{Black Box Outcome Explanation.}  
Given a black box predictor $b$ and an instance $x$, the \textit{black box outcome explanation problem} consists in providing an explanation $e$ for the decision $b(x)=y$. 
We approach the problem by learning an interpretable predictor $c$ that reproduces and accurately mimes the \textit{local} behavior of the black box. 
An explanation of the decision is then derived from $c$.
By \emph{local}, we mean focusing on the behavior of the black box in the neighborhood of the specific instance $x$, without aiming at providing a single description of the logic of the black box for all possible instances. 
The neighborhood of $x$ is not given, but rather it has to be generated as part of the explanation process. 
However, we assume that some knowledge is available about the characteristics of the feature space $\mathcal{X}^{(m)}$, in particular the ranges of admissible values for the domains of features and, possibly, the (empirical) distribution of features. 
Nothing is instead assumed about the process of constructing the black box $b$. 
Let us formalize the problem, and the approach based on interpretable models.

\begin{definition}[Black Box Outcome Explanation]\label{def:problem}
Let $b$ be a black box, and $x$ an instance whose decision $b(x)$ has to be explained. 
The \emph{black box outcome explanation problem} consists in finding an explanation $e \in E$ belonging to a human-interpretable domain $E$. 
\end{definition}

\begin{definition}[Explanation Through Interpretable Models]
\label{def:expinterpretable}
Let $c = \zeta(b, x)$ be an interpretable predictor derived from the black box $b$ and the instance $x$ using some process $\zeta(\cdot,\cdot)$.
An explanation $e \in E$ is obtained through $c$, if $e = \varepsilon(c, x)$ for some explanation logic $\varepsilon(\cdot,\cdot)$ which reasons over $c$ and $x$.
\end{definition}

One point is still missing: which is a comprehensible domain $E$ of explanations? 
We will define an explanation $e$ as a pair of objects:
$$e = \langle r = p \rightarrow y, \Phi \rangle$$
The first component $r = p \rightarrow y$ is a decision rule describing the reason for the decision value $y = c(x)$ 
The second component $\Phi$ is a set of counterfactual rules, namely the minimal number of changes in the feature values of $x$ that would reverse the decision of the predictor.
Let us consider as an example the following explanation for a loan request for user $x = \{ (\mathit{age}{=}22), (\mathit{job} = \mathit{none}), (\mathit{amount}{=}\mathit{10k}), (\mathit{car}{=} \mathit{no})$:
\begin{align*}
e = \langle & r = \{\mathit{age}{\leq}25, \mathit{job}{=}\mathit{none}, \mathit{amount}{>}\mathit{5k}\} {\rightarrow} \mathit{deny},\\ 
& \Phi = \{(\{\mathit{age}{>}25, \mathit{amount}{\leq}5k\} {\rightarrow} \mathit{grant}),\\ 
&\;\;\;\;\;\;\;\;\;(\{\mathit{job}{=}clerk, \mathit{car}{=}\mathit{yes}\}{\rightarrow} \mathit{grant}) 
\} \rangle
\end{align*}
Here, the decision $\mathit{deny}$ is due to the age lower than 25, the absence of job and an amount greater than 5k (see component $r$). For changing the decision instead it is required either an age higher than 25 and a smaller amount, or owning a clerk job and a car (see component $\Phi$).
Details are provided in the rest of the section.

\smallskip
In a \textit{decision rule} (simply, a rule) $r$ of the form $p \rightarrow y$, the decision $y$ is the \emph{consequence} of the rule, while the \textit{premise} $p$ is a boolean condition on feature values. 
We assume that $p$ is the conjunction of split conditions $\mathit{sc}$ of the form $a \in  [v_1, v_2]$, where $a$ is a feature and $v_1, v_2$ are values in the domain of $a$ extended with\footnote{
    Using $\pm \infty$ we can model with a single notation typical univariate split conditions, such as equality ($a = v$ as $a \in [v, v]$), upper bounds ($a \leq v$ as $a \in [-\infty, v]$), strict lower bounds ($a > v$ as $a \in [v+\epsilon, \infty]$ for a sufficiently small $\epsilon$).
    However, since our method is parametric to a decision tree induction algorithm, split conditions can also be multivariate, e..g,~$a \leq b + v$ for $a, b$ features (as in oblique decision trees).} $\pm \infty$.
An instance $x$ \emph{satisfies} $r$, or $r$ \emph{covers} $x$, if the boolean condition $p$ evaluates to true for $x$, i.e.,~if $\mathit{sc}(x)$ is true for every $\mathit{sc} \in p$.
For example, the rule $\{\mathit{age}{\leq}25, \mathit{job}{=}\mathit{none}\} {\rightarrow} \mathit{deny}$ is satisfied by $x_0 = \{ (\mathit{age}{=}22), (\mathit{job}{=}\mathit{none}) \}$ and not satisfied by $x_1 = \{ (\mathit{age}{=}22), (\mathit{job}{=}\mathit{clerk}) \}$.
We say that $r$ \textit{is consistent} with $c$, if $c(x) = y$ for every instance $x$ that satisfies $r$. 
Consistency means that the rule specifies some conditions for which the predictor makes a specific decision. 
When the instance $x$ for which we have to explain the decision satisfies $p$, the rule $p \rightarrow y$ represents \textit{a motivation for taking } a decision value, i.e., $p$ locally explains why $b$ returned $y$.

Consider now a set $\delta$ of split conditions.
We denote the update of $p$ by $\delta$ as $p[\delta] = \delta \cup \{ (a \in [v_1, v_2]) \in p \ |\ \not \exists [w_1, w_2], (a \in [w_1, w_2]) \in \delta \}$. 
Intuitively, $p[\delta]$ is the logical condition $p$ with ranges for attributes overwritten as stated in $\delta$, e.g. $\{\mathit{age} {\leq} 25, \mathit{job}{=}\mathit{none}\} [\mathit{age} {>} 25]$ is $\{\mathit{age} > 25, \mathit{job}{=}\mathit{none}\}$.
A \textit{counterfactual rule} for $p$ is a rule of the form $p[\delta] \rightarrow \hat{y}$, for $\hat{y} \neq y$. We call $\delta$ a \textit{counterfactual}. Consistency is meaningful also for counterfactual rules, meaning that the rule is an instance of the decision logic of $c$. 
A counterfactual $\delta$ describes \textit{what} features to change and \textit{how} to change them to get an outcome different from $y$. Since $c$ predicts either $y$ or $\hat{y}$, if such changes are applied to the given instance $x$, the predictor $c$ will return a different decision. Continuing the example before, changing the age feature of $x_0$ to any value greater than $25$ will change the predicted outcome of $c$ from $\mathit{deny}$ to $\mathit{grant}$.
An expected property of a consistent counterfactual rule $p[\delta] \rightarrow \hat{y}$ is that it should be minimal w.r.t.~$x$.~ Minimality is measured\footnote{Such a measure can be extended to exploit additional knowledge on the feature domains in order not to generate invalid or unrealistic rules. 
E.g.,~the split condition $\mathit{age} \le 25$ appears closer than $\mathit{age} > 30$ for an instance with $\mathit{age} = 26$. However, it is not actionable: an individual cannot lower her age, or change her race or gender.} w.r.t.~the number of split conditions in $p[\delta]$ not satisfied by $x$. Formally, we define $\mathit{nf}(p[\delta], x) = |\{\mathit{sc} \in p[\delta] \ |\ \neg \mathit{sc}(x)\}|$ (where $\mathit{nf}(\cdot,\cdot)$ stands for the number of falsified split conditions), and, when clear from the context, we simply write $\mathit{nf}$.
For example,~$\{\mathit{age} < 25, \mathit{job}{=}\mathit{clerk}\} \rightarrow \mathit{grant}$ is a counterfactual with two conditions falsified by $x_0$. It is not minimal if the counterfactual $\{\mathit{age} {>} 25, \mathit{job}{=}\mathit{none}\} \rightarrow \hat{y}$, with only one falsified condition, is consistent for $c$. In summary, a counterfactual rule $p[\delta] \rightarrow \hat{y}$ is a (minimal) \textit{motivation for reversing} a decision value.

\smallskip
We are now in the position to formally introduce the notion of explanation that we are able to provide.

\begin{definition}[Local Explanation] \label{def:explanation}
Let $x$ be an instance, and $c(x) = y$ be the decision of $c$.
A local explanation $e = \langle r, \Phi \rangle$ is a pair of: a decision rule $r = (p \rightarrow y)$ 
consistent with $c$ and satisfied by $x$; and, a set  $\Phi = \{p[\delta_1] \rightarrow \hat{y}, \dots, p[\delta_v]  \rightarrow \hat{y}\}$ of counterfactual rules for $p$ consistent with $c$.
\end{definition}

This definition completes the elements of the black box outcome explanation problem. 
A solution to the problem will then consists of: 
\emph{(i)} computing an interpretable predictor $c$ for a given black box $b$ and an instance $x$, i.e.,~defining the function $\zeta(\cdot,\cdot)$ according to Definition~\ref{def:expinterpretable}; 
\emph{(ii)} deriving a local explanation from $c$ and $x$, i.e.,~defining the explanation logic $\varepsilon(\cdot,\cdot)$ according to Definition~\ref{def:expinterpretable}.

\section{Proposed Method}
\label{sec:method}

We propose \textbf{LORE} (\textbf{LO}cal \textbf{R}ule-based \textbf{E}xplanations, Algorithm~\ref{alg:lore}) as a solution to the black box outcome explanation problem.
An interpretable predictor $c$ is built for a given black box $b$ and instance $x$ by first generating a set of $N$ neighbor instances of $x$ through a \emph{genetic algorithm}, and then extracting from such a set a \emph{decision tree} $c$. 
A local explanation, consisting of a single rule $r$ and a set of counterfactual rules $\Phi$, is then derived from the structure of $c$.

\begin{algorithm2e}[tb]
	\small
	\caption{$\mathit{LORE}(x, b)$}
	\label{alg:lore}
	\SetKwInOut{Input}{Input}
	\SetKwInOut{Output}{Output}
	\Input{
    $x$ - instance to explain,
    $b$ - black box,
    $N$ - \# of neighbors}
	\Output{$e$ - explanation of $x$}
	\BlankLine
    $G\leftarrow10$; $\mathit{pc}\leftarrow0.5$; $\mathit{pm}\leftarrow0.2$; \hfill\texttt{\scriptsize// init. parameters}\\
    $Z_{=} \leftarrow \mathit{GeneticNeigh}(x, \mathit{fitness}^x_{=}, b, N/2, G, \mathit{pc}, \mathit{pm})$\hfill\texttt{\scriptsize// generate neigh.}\\
    $Z_{\neq} \leftarrow \mathit{GeneticNeigh}(x, \mathit{fitness}^x_{\neq}, b, N/2, G, \mathit{pc}, \mathit{pm})$\hfill\texttt{\scriptsize// generate neigh.}\\
    $Z \leftarrow Z_{=} \cup Z_{\neq}$; \hfill\texttt{\scriptsize// merge neighborhoods}\\
    $c \leftarrow \mathit{BuildTree}(Z)$; \hfill\texttt{\scriptsize// build decision tree}\\
    $r = (p {\rightarrow} y) \leftarrow \mathit{ExtractRule}(c, x)$; \hfill\texttt{\scriptsize// extract decision rule}\\
    $\Phi \leftarrow \mathit{ExtractCounterfactuals}(c, r, x)$; \hfill\texttt{\scriptsize// extract counterfactuals}\\
	\Return{$e = \langle r , \Phi \rangle$}\;
\end{algorithm2e}

{
\setlength{\textfloatsep}{2mm}
\setlength{\intextsep}{2mm}
\setlength{\floatsep}{2mm}

\subsection{Neighborhood Generation}
\label{sec:neigh}
The goal of this phase is to identify a set of instances $Z$, with feature characteristics close to the ones of $x$, that is able to reproduce the local decision behavior of the black box $b$. 
Since the objective is to learn a predictor, the neighborhood should be flexible enough to include instances with both decision values, namely $Z = Z_= \cup  Z_{\neq}$ where instances $z \in Z_=$ are such that $b(z)=b(x)$, and instances $z \in Z_{\neq}$ are such that $b(z) \neq b(x)$. 
In Algorithm~\ref{alg:lore}, we extract balanced subsets $Z_=$ and $Z_{\neq}$ (lines 2--3), and then put $Z = Z_= \cup  Z_{\neq}$ (line 4).
This task differs from approaches to instance \textit{selection} \cite{DBLP:journals/air/Olvera-LopezCTK10}, based on genetic algorithms \cite{DBLP:journals/kbs/TsaiEC13} (also specialized for decision trees \cite{geneticselection2009}), in that their objective is to select a subset of instances from an available training set. 
In our case, instead we cannot assume that the training set used to train $b$ is available, or not even that $b$ is a supervised machine learning predictor for which a training set exists. 
Our task is instead similar to instance \textit{generation} in the field of active learning \cite{DBLP:journals/kais/FuZL13}, also including evolutionary approaches \cite{DBLP:journals/ijamc-igi/DerracGH10}.
We adopt an approach based on a \emph{genetic algorithm} which generates $z \in Z_= \cup  Z_{\neq}$ by maximizing the following fitness functions:
\[  
\begin{array}{l}
      \mathit{fitness}^x_{=}(z) = I_{b(x)=b(z)} + (1-d(x,z)) - I_{x=z}
\\
      [2mm]
      \mathit{fitness}^x_{\neq}(z) = I_{b(x)\neq b(z)} + (1-d(x,z)) - I_{x=z}\\ 
\end{array} 
\]
where $d: \mathcal{X}^{(m)} \rightarrow [0, 1]$ is a distance function, $I_{\mathit{true}} = 1$, and  $I_{\mathit{false}} = 0$.
The first fitness function looks for instances $z$ similar to $x$ (term $1-d(x,z)$), but not equal to $x$ (term $I_{x=z}$) for which the black box $b$ produces the same outcome as $x$ (term $I_{b(x)=b(z)}$). The second one leads to the generation of instances $z$ similar to $x$, but not equal to it, for which $b$ returns a different decision. 
Intuitively, for an instance $z_0$ such that $b(x) \neq b(z_0)$ and $x \neq z_0$, it turns out $\mathit{fitness}^x_{=}(z_0) < 1$. 
For any instance $z_0$ such that $b(x) = b(z_0)$, instead, we have 
$\mathit{fitness}^x_{=}(z_0) \geq 1$. Finally,  $\mathit{fitness}^x_{=}(x) = 1$. 
Thus, maximization of $\mathit{fitness}^x_{=}(\cdot)$ occurs necessarily for instances different from $x$ and whose prediction is equal to $b(x)$.

\begin{algorithm2e}[tb]
	\small
	\caption{$\mathit{GeneticNeigh}(x, \mathit{fitness}, b, N, G, \mathit{pc}, \mathit{pm})$}
	\label{alg:gp}
	\SetKwInOut{Input}{Input}
	\SetKwInOut{Output}{Output}
	\Input{
    $x$ - instance to explain,
    $b$ - black box,
    $\mathit{fitness}$ -  fitness function,
    $N$ - population size,
    $G$ - \# of generations,
    $\mathit{pc}$ - crossover probability, 
    $\mathit{pm}$ - mutation probability}
	\Output{$Z$ - neighbors of $x$}
	\BlankLine
	$P_0 \leftarrow \{ x | \forall 1 \dots N \}$; $i \leftarrow 0$; \hfill\texttt{\scriptsize// population init.}\\
    $\mathit{evaluate}(P_0, \mathit{fitness}, b)$; \hfill\texttt{\scriptsize// evaluate population}\\
    \While{$i < G$}{
    	$P_{i+1} \leftarrow \mathit{select}(P_i)$; \hfill\texttt{\scriptsize// select sub-population}\\
        $P'_{i+1} \leftarrow \mathit{crossover}(P_{i+1}, \mathit{pc})$; \hfill\texttt{\scriptsize// mix records}\\
        $P''_{i+1} \leftarrow \mathit{mutate}(P'_{i+1}, \mathit{pm})$; \hfill\texttt{\scriptsize// perform mutations}\\
        $\mathit{evaluate}(P''_{i+1}, \mathit{fitness}, b)$; \hfill\texttt{\scriptsize// evaluate population}\\
        $P_{i+1} = P''_{i+1}$; $i \leftarrow i + 1$ \hfill\texttt{\scriptsize// update population}\\
    }
    $Z \leftarrow P_i$
	\Return{$Z$}\;
\end{algorithm2e}

Like neural networks, genetic algorithms \cite{holland1992adaptation} are based on the biological metaphor of evolution. 
They have three distinct aspects.
\emph{(i)} The potential solutions of the problem are encoded into representations that support the \textit{variation} and \textit{selection} operations. 
In our case these representations, generally called chromosomes, correspond to instances in the feature space $\mathcal{X}^m.$
\emph{(ii)} A fitness function evaluates which chromosomes are the ``best life forms'', that is, most appropriate for the  result. 
These are then favored in \emph{survival} and \emph{reproduction}, thus shaping the next generation according to the fitness function.
In our case, these instances correspond to those similar to $x$, according to distance $d(\cdot, \cdot)$, and with the same/different outcome returned by the black box $b$ depending on the fitness function $\mathit{fitness}^x_=$ or $\mathit{fitness}^x_{\neq}$.
\emph{(iii)} Mating (called crossover) and mutation produce a new generation of chromosomes by recombining features of their parents. 
The final generation of chromosomes, according to a stopping criterion, is the one that best fit the solution. 

}

Algorithm~\ref{alg:gp} generates the neighborhoods $Z_=$ and $Z_{\neq}$ of $x$ by instantiating the evolutionary approach of \cite{back2000evolutionary}. 
Using the terminology of \cite{DBLP:journals/ijamc-igi/DerracGH10}, it is an instance of generational genetic algorithms for evolutionary prototype generation. 
However, prototypes are a condensed subset of a training set that enable optimization in predictor learning. We aim instead at generating new instances that separate well the decision boundary of the black box $b$.
The usage of classifiers within fitness functions of genetic algorithms can be found in \cite{eshelman1991chc,baluja1994population,wu2006optimal,cano2005stratification}. 
However, the classifier they use is always the one for which the population must be selected or generated for and not another one (the black box) like in our case.
Algorithm~\ref{alg:gp} first initializes the population $P_0$ with $N$ copies of the instance $x$ to explain.
Then it enters the evolution loop that begins by \textit{selection} of the $P_{i+1}$ population having the highest fitness score. 
After that, the crossover operator is applied on a proportion of $P_{i+1}$ according to the $\mathit{pc}$ probability, the resulting and the untouched individuals are placed in $P'_{i+1}$. 
We use a \emph{two-point crossover} which selects two parents and two crossover features at random, and then swap the crossover feature values of the parents (see Figure~\ref{tab:crossover}).
Thereafter, a proportion of $P'_{i+1}$, determined by $\mathit{pm}$, is mutated and placed in $P''_{i+1}$. 
The unmutated individuals are also added to $P''_{i+1}$. 
Mutation consists of replacing features values at random according to the empirical distribution\footnote{In experiments, we derive such a distribution from the test set of instances to explain.} of a feature  (see Figure~\ref{tab:mutation}). 
Individuals in $P''_{i+1}$ are evaluated according to the fitness function, and the evolution loop continues until $G$ generations are completed.
Finally, the best individuals according to the fitness function are returned.  Algorithm~\ref{alg:gp} is run twice, once using the fitness function $\mathit{fitness}^x_{=}$ to derive neighborhood instances $Z_{=}$ with the same decision as $x$, and once using $\mathit{fitness}^x_{\neq}$ to derive neighborhood instances $Z_{\neq}$ with different decision as $x$. 
Finally, we set $Z = Z_= \cup Z_{\neq}$.

\begin{figure}[tb]
\centering
\footnotesize
\begin{minipage}[b]{0.48\linewidth}
    \centering
    \setlength{\tabcolsep}{1.5mm}
    \begin{tabular}{cc!{\color{green}\vrule width 1pt }c|c!{\color{green}\vrule width 1pt}c}
     \cline{2-5} 
    \multicolumn{1}{c|}{parent 1} & \cellcolor[HTML]{c0a6c6}25 & \cellcolor[HTML]{c0a6c6}clerk & \cellcolor[HTML]{c0a6c6}10k & \multicolumn{1}{c|}{\cellcolor[HTML]{c0a6c6}yes} \\ \cline{2-5} 
    \cline{2-5} 
    \multicolumn{1}{c|}{parent 2} & \cellcolor[HTML]{FFFFFF}30 & \cellcolor[HTML]{FFFFFF}other & \cellcolor[HTML]{FFFFFF}5k & \multicolumn{1}{c|}{\cellcolor[HTML]{FFFFFF}no} \\ \cline{2-5} 
     &  &  \multicolumn{2}{c!{\color{green}\vrule width 1pt}}{$\downarrow$}  &  \\ \cline{2-5}
    \multicolumn{1}{c|}{children 1} & \cellcolor[HTML]{c0a6c6}25 & other & 5k & \multicolumn{1}{c|}{\cellcolor[HTML]{c0a6c6}yes} \\ \cline{2-5} 
    \multicolumn{1}{c|}{children 2} & 30 & \cellcolor[HTML]{c0a6c6}clerk & \cellcolor[HTML]{c0a6c6}10k & \multicolumn{1}{c|}{no} \\ \cline{2-5} 
    \end{tabular}
    \caption{Crossover.}
    \label{tab:crossover}
\end{minipage}\hfill
\begin{minipage}[b]{0.48\linewidth}
    \centering
    \setlength{\tabcolsep}{1.5mm}
    \begin{tabular}{ccccc}
    \\ \cline{2-5} 
    \multicolumn{1}{l|}{parent} & \multicolumn{1}{l|}{25} & \multicolumn{1}{l|}{clerk} & \multicolumn{1}{l|}{10k} & \multicolumn{1}{l|}{yes} \\ \cline{2-5} 
     & $\downarrow$ &  & $\downarrow$  &  \\ \cline{2-5} 
    \multicolumn{1}{l|}{children} & \multicolumn{1}{l|}{\cellcolor[HTML]{c0a6c6}{\color[HTML]{333333} 27}} & \multicolumn{1}{l|}{clerk} & \multicolumn{1}{l|}{\cellcolor[HTML]{c0a6c6}7k} & \multicolumn{1}{l|}{yes} \\ \cline{2-5} 
    \end{tabular}
    \vspace{-3mm}
    \caption{Mutation}
    \label{tab:mutation}
\end{minipage}
\vspace{-5mm}
\end{figure}

\begin{figure}[t]
    \centering
    \hspace{-2mm}
   		\includegraphics[trim = 2mm 0mm 1mm 0mm, clip,width=0.5\linewidth]{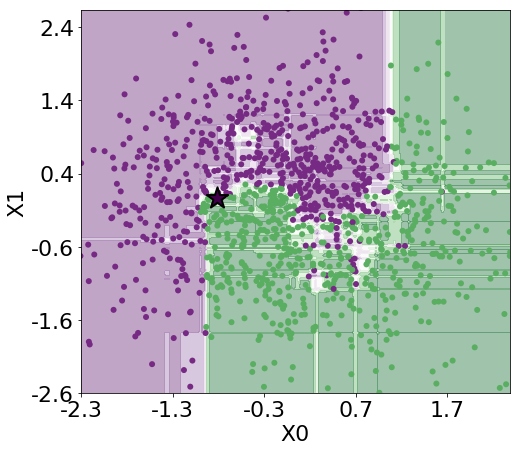}
        \includegraphics[trim = 2mm 0mm 1mm 0mm, clip,width=0.5\linewidth]{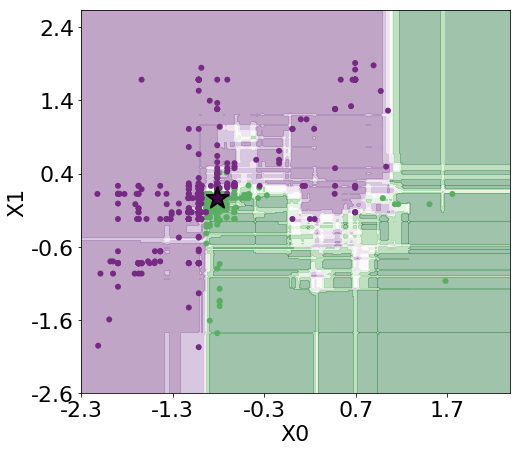} \\
        \includegraphics[trim = 2mm 0mm 1mm 0mm, clip,width=0.5\linewidth]{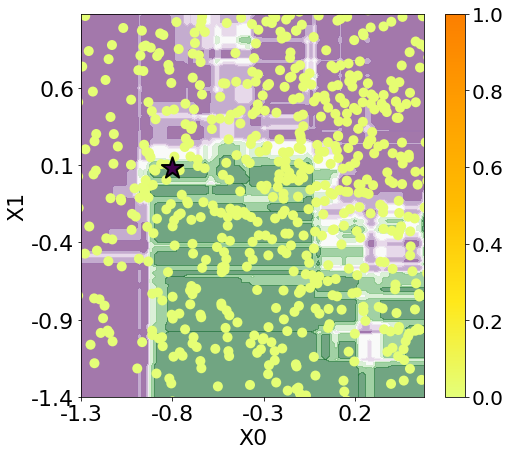}
        \includegraphics[trim = 2mm 0mm 1mm 0mm, clip,width=0.5\linewidth]{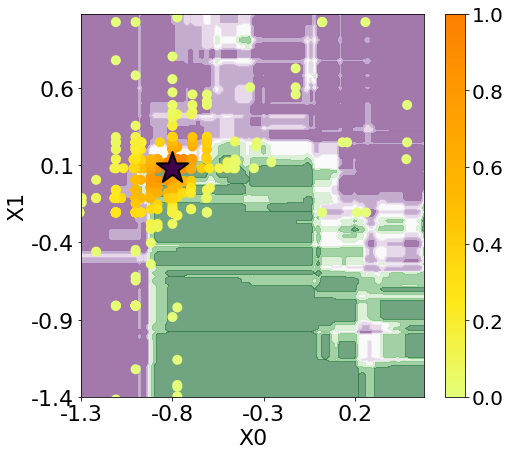} 
\vspace{-5mm}
\caption{
Random forest black box: purple vs green decision. Starred instance $x$. 
\textit{(Top)} Uniformly random  (left) and genetic generation (right).
\textit{(Bottom)} Density of random (left) and genetic (right) generation.
(Best view in color).}
\label{fig:neigh}
\vspace{-2mm}
\end{figure}

Figure~\ref{fig:neigh} shows an example of neighborhood generation for a black box consisting of a random forest model and a bi-dimensional feature space. 
The figure contrasts uniform random generation around a specific instance $x$ (starred) to our genetic approach. The latter yields a neighborhood that is denser in the boundary region of the predictor. 
The density of generated instances will be a key factor in extracting local (interpretable) predictors.

\paragraph{Distance Function}
A key element in the definition of the fitness functions is the distance $d(x, z)$. 
We account for the presence of mixed types of features by a weighted sum of simple matching coefficient for categorical features, and of the  normalized Euclidean distance\footnote{\href{http://reference.wolfram.com/language/ref/NormalizedSquaredEuclideanDistance.html}{reference.wolfram.com/language/ref/NormalizedSquaredEuclideanDistance.html}} for continuous features. 
Formally, assuming $h$ categorical features and $m-h$ continuous ones, we use:
$$d(x, z) = \frac{h}{m} \cdot \mathit{SimpleMatch}(x, z) + \frac{m-h}{m} \cdot \mathit{NormEuclid}(x, z).$$
Our approach is parametric to $d(\cdot, \cdot)$, and it can readily be applied to improved heterogeneous distance functions
\cite{DBLP:journals/prl/McCaneA08}. 
Empirical results with different distance functions are reported in Section \ref{sect:distance}.

\subsection{Local Rule-Based Classifier and Explanation Extraction}
\label{sec:local_classifier}

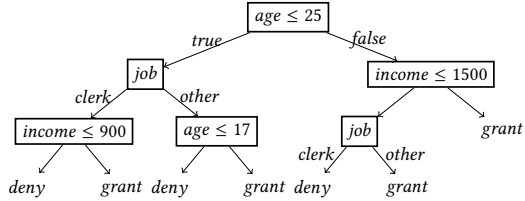
\begin{figure}
\begin{center}
\scalebox{0.8}{
\begin{tikzpicture}[->,
level/.style = {sibling distance = 15em/#1, level distance = 3em},
split/.style = {shape=rectangle, draw, align=center, thick, scale=1.0},
leaf/.style =  {draw=none,fill=none, align=center} 
edge from parent/.style = {->, draw, line width=5.0}]
\node[split](root){$\mathit{age}\leq25$}
child{ node[split](c1){$\mathit{job}$} 
	child{ node[split](c3){$\mathit{income}\leq900$} 
		child{ node[leaf]{$\mathit{deny}$} }
        child{ node[leaf]{$\mathit{grant}$} }
    }
    child{ node[split](c4){$\mathit{age}\leq17$} 
    	 child{ node[leaf]{$\mathit{deny}$} }
        child{ node[leaf]{$\mathit{grant}$} }
    }
} 	child{ node[split](c2){$\mathit{income}\leq1500$}
    	child{ node[split](c5) {$\mathit{job}$}
        	child{ node[leaf](c6){$\mathit{deny}$} }
        	child{ node[leaf](c7){$\mathit{grant}$} }}
        child{ node[leaf]{$\mathit{grant}$} }
};
\path (root) -- (c1) node [near start, left]  {$\mathit{true}$};
\path (root) -- (c2) node [near start, right]  {$\mathit{false}$};
\path (c1) -- (c3) node [near start, left]  {$\mathit{clerk}$};
\path (c1) -- (c4) node [near start, right]  {$\mathit{other}$};
\path (c5) -- (c6) node [near start, left]  {$\mathit{clerk}$};
\path (c5) -- (c7) node [near start, right]  {$\mathit{other}$};
\end{tikzpicture}
}
\end{center}
\vspace{-3ex}
\caption{Example decision tree.\label{ex:dec}}
\vspace{-5mm}
\end{figure}

Given the neighborhood $Z$ of $x$, the second step is to build an interpretable predictor $c$ trained on the instances $z \in Z$ labeled with the black box decision $b(z)$. Such a predictor is intended to mimic the behavior of $b$ locally in the $Z$ neighborhood. Also, $c$ must be interpretable, so that an explanation for $x$ (decision rule and counterfactuals) can be extracted from it. The \textbf{LORE} approach considers decision tree classifiers due to the following reasons: \emph{(i)}  decision rules can  naturally be derived from a root-leaf path in a decision tree; and, \emph{(ii)} counterfactuals can be extracted by symbolic reasoning over a decision tree. For a decision tree $c$, we derive an explanation $e = \langle r,  \Phi \rangle$ as follows.
The decision rule $r = (p \rightarrow y)$ is formed by including in $p$ the split conditions on the path from the root to the leaf node that is satisfied by the instance $x$, and setting $y = c(x)$. By construction, the rule $r$ is consistent with $c$ and satisfied by $x$. Consider now the counterfactual rules in $\Phi$. 
Algorithm~\ref{alg:extract_counter} 
looks for all paths in the decision tree $c$ leading to a decision $\hat{y} \neq y$. Fix one of such paths, and let $q$ be the 
conjunction of split conditions in it. Again by construction, $q \rightarrow \hat{y}$ is a counterfactual rule consistent with $c$. Notice that the counterfactual $\delta$ for which $q = p[\delta]$ has not to be explicitly computed\footnote{However, it can be done as follows. Consider the path from the leaf of $p$ to the leaf of $q$. When moving from a child to a father node, we retract the split condition. 
E.g., $a \leq v_2$ is retracted from $\{ a \in [v_1, v_2] \}$ by adding $a \in [v_1, +\infty]$ to $\delta$. When moving from a father node to a child, we add the split condition to $\delta$.} -- this is a benefit of using decision trees. 
Among all such $q$'s, only the ones with minimum number of split conditions $\mathit{sc}$ not satisfied by $x$ (line 4 of Algorithm~\ref{alg:extract_counter}) are kept in $\Phi$. 
As an example, consider the decision tree in Figure~\ref{ex:dec}, and the instance $x = \{ (\mathit{age}, 22), (\mathit{job}, \mathit{clerk}), (\mathit{income}, 800)\}$ for which the decision $\mathit{deny}$ (e.g.,~of a loan) has to be explained. The path followed by $x$ is the leftmost one in the tree. The decision rule extracted from the path is $\{ \mathit{age} \leq 25, \mathit{job}{=}\mathit{clerk}, \mathit{income} \leq 900 \} \rightarrow \mathit{deny}$. 
There are four paths leading to the opposite decision: $q_1$ $=$ $\{ \mathit{age} \leq 25, \mathit{job}{=}\mathit{clerk}, \mathit{income} > 900  \}$, $q_2$ $=$ $\{ 17 < \mathit{age} \leq 25, \mathit{job} = \mathit{other}\}$, $q_3$ $=$ $\{ \mathit{age} > 25, \mathit{income} \leq 1500, \mathit{job} = \mathit{other}\}$, and $q_4$ $=$ $\{ \mathit{age} > 25, \mathit{income} > 1500\}$. It turns out: $\mathit{nf}(q_1, x)=1$, $\mathit{nf}(q_2, x)=1$, $\mathit{nf}(q_3, x)=2$, and $\mathit{nf}(q_4, x)=2$. \mbox{Thus, $\Phi = \{ q_1 {\rightarrow} \mathit{deny}, q_2 {\rightarrow} \mathit{deny} \}$.}

\begin{algorithm2e}[t]
	\small
\caption{$\mathit{ExtractCounterfactuals}(c, r, x)$}
	\label{alg:extract_counter}
	\SetKwInOut{Input}{Input}
	\SetKwInOut{Output}{Output}
	\Input{
    $c$ - decision tree,
    $r$ - rule ($p \rightarrow y)$,
    $x$ - instance to explain}
	\Output{$\Phi$ - set of counterfactual rules for $p$}
	\BlankLine
    $Q \leftarrow \mathit{GetPathsWithDifferentLabel}(c, y)$; \hfill\texttt{\scriptsize// get paths with label $\hat{y} \neq y$}\\
    $\Phi \leftarrow \emptyset$; $\mathit{min} \leftarrow +\infty$; \hfill\texttt{\scriptsize// init. counterfactual set}\\
    \For{$q \in Q$}{
    	$\mathit{qlen} \leftarrow \mathit{nf}(q, x) = |\{ \mathit{sc} \in q \ |\ \neg \mathit{sc}(x) \}|$\\
    	\lIf{$\mathit{qlen} < \mathit{min} $} {
        	$\Phi \leftarrow \{ q \rightarrow \hat{y} \}$; $\mathit{min}  \leftarrow \mathit{qlen}$} \lElseIf{$\mathit{qlen} = \mathit{min} $}{
			$\Phi \leftarrow \Phi \cup \{ q \rightarrow \hat{y} \}$}
    }
	\Return{$\Phi$}\;
\end{algorithm2e}

As a further output, \textbf{LORE} computes a \textit{counterfactual instance} starting from a counterfactual rule $q \rightarrow \hat{y}$ and $x$. Among all possible instances that satisfy $q$, we choose the one that minimally changes attributes from $x$ according to $q$. This is done by looking at the split conditions falsified by $x$: $\{\mathit{sc} \in q \ |\ \neg \mathit{sc}(x) \}$, and modifying the lower/upper bound in $\mathit{sc}$ that is closer to the value in $x$. As an example, for the above path $q_1$, the counterfactual instance of $x$ is $\{ (\mathit{age}, 22), (\mathit{job}, \mathit{clerk}), (\mathit{income}, 900+\epsilon)\}$, and for $q_2$ is $\{ (\mathit{age}, 22), (\mathit{job}, \mathit{other}), (\mathit{income}, 800)\}$.



\section{Experiments}
\label{sec:experiments}
\textbf{LORE} has been developed in Python\footnote{The source code and datasets will be available at \textit{anonimized url}. 
The experiments were performed on Ubuntu 16.04.1 LTS 64 bit, 32 GB RAM, 3.30GHz Intel Core i7}, using, for the genetic neighborhood generation, the \texttt{deap} library~\cite{DEAP_JMLR2012}, 
and for decision tree induction (the interpretable predictor), the \texttt{yadt} system~\cite{ruggieri2004yadt}, which is a C4.5 implementation with multi-way splits of categorical attributes. 
After presenting the experimental setup, we report next: \emph{(i)} some analyses on the effect of the genetic algorithm parameters for the neighborhood generation; \emph{(ii)} evidence that the local genetic neighborhood is more effective than a global approach; 
\emph{(iii)} a qualitative and quantitative comparison with na\"ive baselines and state of the art competitors.

\subsection{Experimental Setup}

We ran experiments on three real-world \emph{tabular} datasets: 
\emph{adult}, \emph{compas} and \emph{german}\footnote{\url{https://archive.ics.uci.edu/ml/datasets/adult}, \url{https://github.com/propublica/compas-analysis}, \url{https://archive.ics.uci.edu/ml/datasets/statlog+(german+credit+data)}}.
In each of them, an instance represents attributes of an individual person. 
All datasets includes both categorical and continuous features.

The \emph{adult} dataset from UCI Machine Learning Repository, includes $48,842$ instances with demographic information like age, workclass, marital-status, race, capital-loss, capital-gain etc.
The income divides the population into two classes ``<=50K'' and ``>50K''.

The \emph{compas} dataset from ProPublica contains the features used by the COMPAS algorithm for scoring defendants and their risk (Low, Medium and High), for over $10,000$ individuals. 
We considered two classes ``Low-Medium'' and ``High'' risk, and we use the following features: age, sex, race, priors\_count, days\_b\_scree\\ning\_arrest, length\_of\_stay, c\_charge\_degree, is\_recid. 
 
In the \emph{german} dataset from UCI Machine Learning Repository each person of the $1,000$ entries is classified as a ``good'' or ``bad'' creditor according to attributes like age, sex, checking\_account, credit\_amount, duration, purpose, etc.        

We experimented the following predictors as black boxes: support vector machines with RBF kernel (\textbf{SVM}), random forests with 100 trees (\textbf{RF}), and multi-layer neural networks with `lbfgs' solver (\textbf{NN}).
Implementations of the predictors are from the \texttt{scikit-learn} library. Unless differently stated, default parameters were used for both the black boxes and the libraries of \textbf{LORE}. Missing values were replaced by the mean for continuous features and by the mode for categorical ones.

Each dataset was randomly split into train (80\% instances), and test (20\% instances). 
The former is used to train black box predictors. 
The latter, denoted by $X$, is the set of instances for which the black box decision have to be explained. 
In the following, for some fixed set of instances, we denote by $Y$ the set of decisions provided by the interpretable predictor $c$ and by $\hat{Y}$ the decisions provided by the black box $b$ on the same set.


We consider the following properties in evaluating the mimic performances of the decision tree $c$ inferred by \textbf{LORE} and of the explanations returned by it against the black box classifier $b$:
\begin{itemize}
\item $\mathit{\textbf{\textit{fidelity}}}(Y, \hat{Y}) \in [0, 1]$. It compares the predictions of  $c$ and of the black box $b$ on the instances $Z$ used to train $c$ \cite{doshi2017towards}. It answer the question: how good is $c$ at mimicking $b$?
\item $\mathit{\textbf{\textit{l-fidelity}}}(Y, \hat{Y}) \in [0, 1]$. It compares the predictions of $c$ and $b$ on the instances $Z_x$ covered by the decision rule in a local (hence ``l-'') explanation for $x$.  
It answers the question: how good is the decision rule at mimicking $b$?
\item $\mathit{\textbf{\textit{cl-fidelity}}}(Y, \hat{Y}) \in [0, 1]$.  It compares the predictions of $c$ and $b$ on the instances $Z_x$ covered by the counterfactual rules in a local explanation for $x$.
\item $\mathit{\textbf{\textit{hit}}}(y, \hat{y}) \in \{0, 1\}$. It compares the predictions of $c$ and $b$ on the instance $x$ under analysis. It returns $1$ if $y = c(x)$ is equal to $\hat{y} = b(x)$, and $0$ otherwise. 
\item $\mathit{\textbf{\textit{c-hit}}}(y, \hat{y}) \in \{0, 1\}$. It compares the predictions of $c$ and $b$ on a counterfactual instance of $x$ built from counterfactual rules in a local explanation of $x$.
\end{itemize}
We measure the first three of them by the f1-measure \cite{tan2005introduction}. Aggregated values of f1 and hit/c-hit are reported by averaging them over the the set of test instances $x \in X$.


\begin{figure}[t]
    \centering
    \hspace{-2mm}
       		\includegraphics[trim = 2mm 0mm 1mm 0mm, clip,width=0.49\linewidth]{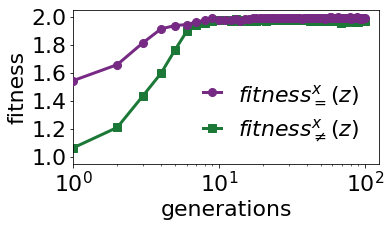}
            \includegraphics[trim = 2mm 0mm 1mm 0mm, clip,width=0.49\linewidth]{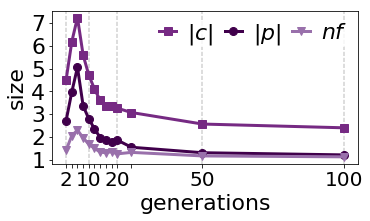} 
                \includegraphics[trim = 2mm 0mm 1mm 0mm, clip,width=0.49\linewidth]{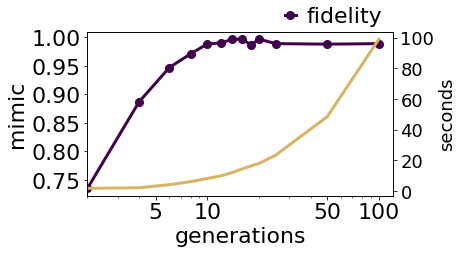} 
   		\includegraphics[trim = 2mm 0mm 1mm 0mm, clip,width=0.49\linewidth]{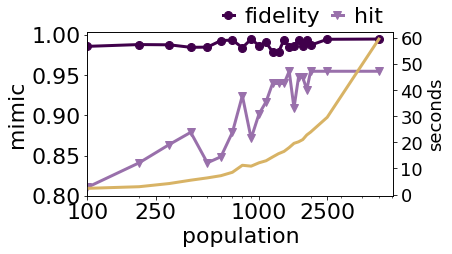} 
 	\vspace{-2mm}
\caption{Impact of the number of generations $G$ and of population size $N$ parameters of the genetic neighborhood generation. Bottom plots also report elapsed running times.
}
\label{fig:gpparameters}
\vspace{-1mm}
\end{figure}

\begin{figure*}[t]
    \centering
    \hspace{-2mm}
    	\includegraphics[trim = 2mm 0mm 1mm 0mm, clip,width=0.33\linewidth]{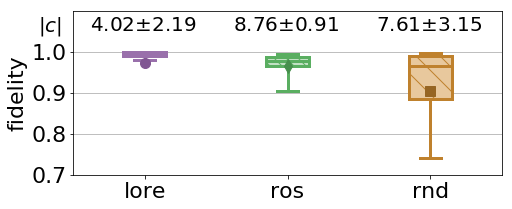} 
   		\includegraphics[trim = 2mm 0mm 1mm 0mm, clip,width=0.33\linewidth]{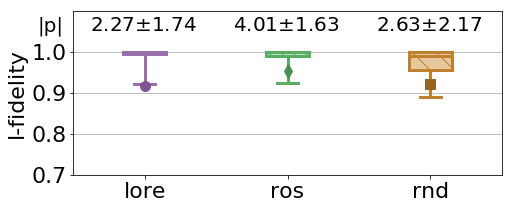}
        \includegraphics[trim = 2mm 0mm 1mm 0mm, clip,width=0.33\linewidth]{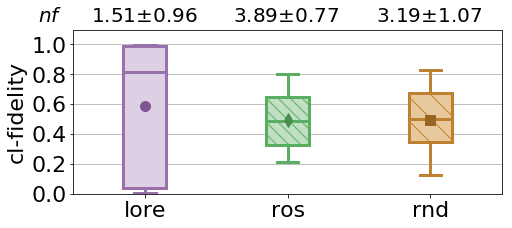}
 	\vspace{-3mm}
\caption{Comparison of neighborhood generations methods.
}
\label{fig:neigh_comp}
\vspace{-1mm}
\end{figure*}

\subsection{Analysis of Neighborhood Generation}
We analyze here the impact of the number of generations $G$ and size of neighborhood $N$ on the performances of instance generation and on the size complexity of the \textbf{LORE} output. We report only results for \emph{german} dataset, since we get similar results for the other ones. The other parameters of Algorithm~\ref{alg:gp} (probabilities of crossover $\mathit{pc}$ and mutation $\mathit{pm}$) are set with the default values of $0.5$ and $0.2$ respectively \cite{back2000evolutionary}.
Figure~\ref{fig:gpparameters} shows in the top plots the value of fitness functions and measures of sizes of local classifier $c$ (decision tree depth), of decision rule (size of the antecedent $p$), and of counterfactual rules (number $\mathit{nf}$ of falsified split conditions). The bottom plots show fidelity (f1-measure) and hit (rate) as well as running times of neighborhood generation.
Fixed $N=1000$, after $10$ generations, the fitness function converges around the optimal value (top left), fidelity is almost maximized (bottom left), and also the measures of sizes (top right) become stable and small.
We then set $G\text{=}10$ in all other experiments.
Figure~\ref{fig:gpparameters}-(bottom right) shows instead that the size $N$ of the neighborhood instances to be generated is relevant for the $\mathit{hit}$ rate but not for $\mathit{fidelity}$.
By taking into account also the running time (right side scale of the bottom plots), a good trade-off is obtained by setting $N\text{=}1000$.

\begin{table}[t]
\setlength{\tabcolsep}{0.7mm}
\small
\begin{tabular}{c|c|c|c|c|c|c}
\toprule
\textit{Distance} & $\mathit{hit}$ & $\mathit{fidelity}$ & $\mathit{l{\text -}fidelity}$ & $|c|$ & $|p|$ & $\mathit{nf}$ \\
\midrule
cosine   & $.938 {\pm} .24$ & $\mathbf{.976 {\pm} .11}$ & $.936 {\pm} .24$ & $4.4 {\pm} 2.5$ & $\mathbf{2.1 {\pm} 1.8}$ & $1.9 {\pm} 1.0$ \\
minmax & $.958 {\pm} .19$ & $.965 {\pm} .15$ & $.956 {\pm} .17$ & $4.5 {\pm} 2.7$ & $2.3 {\pm} 2.3$ & $\mathbf{1.8 {\pm} 0.9}$ \\
neuclid  & $\mathbf{.966 {\pm} .17}$ & $.967 {\pm} .15$ & $.\mathbf{963 {\pm} .19}$ & $\mathbf{4.3 {\pm} 2.6}$ & $2.2 {\pm} 2.1$ & $1.8 {\pm} 1.0$ \\
\bottomrule
\end{tabular}
\caption{Comparison of distance measures.}
\label{tab:dist_comp}
\vspace{-5mm}
\end{table}

\subsection{Comparing Distance Functions}\label{sect:distance}

A key element of the neighborhood generation is the distance function used by the genetic algorithm. 
A legitimate question is whether the results of the approach are affected by the choice of the distance function adopted (see Section~\ref{sec:neigh}). 
For instance, \cite{wachter2017counterfactual} presents considerable differences in their output of counterfactual instance varying the choice of the distance in their stochastic optimization approach.
Table~\ref{tab:dist_comp} reports basic measures contrasting the \textit{normalized Euclidean} distance adopted by \textbf{LORE} with \textit{cosine} and \textit{min-max} distance on \emph{german} dataset.
The table does not highlight any considerable difference. 
This can be justified by the fact that, following instance generation, there are phases, such as decision tree building, that abstract instances to patterns, resulting in resilience against variability due to the distance function adopted.

\subsection{Validation of Local Explanations}
We now compare our local approach with a global approach, and discuss alternative neighborhood instance generation methods.

\begin{table}[t]
\setlength{\tabcolsep}{0.9mm}
\small
\begin{tabular}{c|c|c|c|c|c}
\toprule
\textit{Dataset} & \textit{Method} & $\mathit{hit}$ & $\mathit{fidelity}$ & $\mathit{l{\text -}fidelity}$ & \emph{tree depth} \\
\midrule
{\multirow{2}{*}{adult}} & lore & \textbf{.912 $\pm$ .29} & \textbf{.959 $\pm$ .17} & \textbf{.892 $\pm$ .29} & \textbf{4.16 $\pm$ 0.21} \\
					    & global & .901 $\pm$ .28 & .750 $\pm$ .00 & .873 $\pm$ .27 & 12.00 $\pm$ 0.00 \\
\hline
{\multirow{2}{*}{compas}} & lore & \textbf{.942 $\pm$ .23} & \textbf{.992 $\pm$ .03} & \textbf{.937 $\pm$ .23} & \textbf{4.72 $\pm$ 2.15} \\
                         & global & .902 $\pm$ .29 & .935 $\pm$ .00 & .857 $\pm$ .29 & 12.00 $\pm$ 0.00 \\
\hline
{\multirow{2}{*}{german}} & lore & \textbf{.925 $\pm$ .26} & \textbf{.988 $\pm$ .07} & \textbf{.920 $\pm$ .26} & \textbf{4.95 $\pm$ 2.54} \\
						 & global & .880 $\pm$ .32 & .571 $\pm$ .00 & .824 $\pm$ .31 & 6.00 $\pm$ 0.00 \\
\bottomrule
\end{tabular}
\caption{Local vs global approach.}
\label{tab:local_vs_global}
\vspace{-5mm}
\end{table}

\subsubsection*{Local vs Global Explanations}

Extracting a predictor from the neighborhood of an instance is a winning strategy, if contrasted to an approach that builds a single predictor from all instances in the test set, i.e.,~$Z = X$. In particular, this means that the interpretable predictor will be the same for all instances in the test set. Let compare our approach with such a \texttt{global} approach.
Table~\ref{tab:local_vs_global} reports the mean and standard deviation values of $\mathit{hit}$, $\mathit{fidelity}$, $\mathit{fairness}$ and \emph{tree depth} for each dataset aggregating over the results of the various black boxes. 
While for $\mathit{hit}$ both \textbf{LORE} and \texttt{global} obtain  similar high performances, for the other scores \textbf{LORE}  considerably overtakes \texttt{global}. In particular, the size and depth of the decision tree of the global approach may lead to explanations (decision rules and counterfactuals) more complex to understand than those returned by the proposed local approach \textbf{LORE}.  

\subsubsection*{Comparing Neighborhood Generations}

After concluding that ``local is better than global", we now show that our genetic programming approach improves over the following baselines in the generation of neighborhoods:
\begin{itemize}
\item \texttt{crn} returns as $Z$ the $k=100$ instances from $X$ (the test set) that are \textit{closest} to $x$; 
\item \texttt{rnd} augment the output of \texttt{crn} with additional randomly generated instances so that a stratified $Z$ is obtained; 
\item \texttt{ris} starting from the output of \texttt{rnd} performs the instance selection procedure\footnote{\url{http://contrib.scikit-learn.org/imbalanced-learn}} \emph{CNN} \cite{DBLP:journals/air/Olvera-LopezCTK10};
\item \texttt{ros} starting from the output of \texttt{rnd} performs a random oversampling to balance the decision outcomes in $Z$. 
\end{itemize}

\begin{table}[t]
\setlength{\tabcolsep}{0.7mm}
\small
\begin{tabular}{c|c|c|c|c|c}
\toprule
\textit{Method} & $\mathit{hit}$ &  $\mathit{fidelity}$ &  $\mathit{l{\text -}fidelity}$ & $\mathit{c\text{-}hit}$ & $\mathit{cl\text{-}fidelity}$ \\
\midrule
lore & .962 $\pm$ .19 & \textbf{.993 $\pm$ .04} & \textbf{.959 $\pm$ .19} & \textbf{.588 $\pm$ .42} & \textbf{.756 $\pm$ .40}\\
crd & .924 $\pm$ .26 & .855 $\pm$ .23 & .894 $\pm$ .25 & .349 $\pm$ .26 & .583 $\pm$ .48\\
rnd & .946 $\pm$ .22 & .904 $\pm$ .15 & .920 $\pm$ .22 & .494 $\pm$ .24 & .712 $\pm$ .40\\
ris & .916 $\pm$ .27 & .869 $\pm$ .05 & .870 $\pm$ .26 & .501 $\pm$ .22 & .708 $\pm$ .39\\
ros & \textbf{.968 $\pm$ .17} & .965 $\pm$ .03 & .953 $\pm$ .17 & .491 $\pm$ .22 & .733 $\pm$ .34\\
\bottomrule
\end{tabular}
\caption{Comparison of neighborhood generations methods.
}
\label{tab:neigh_comp}
\vspace{-5mm}
\end{table}

Table~\ref{tab:neigh_comp} reports the aggregated evaluation measures over the various black boxes and datasets. 
\textbf{LORE}  overtook the performance of all the other neighbors generators.
Intuitively, this means that \textbf{LORE}'s genertic programming approach  contributes more than the other methods in capturing/explaining the behavior of the black box, both for direct and counterfactual decisions.
Such a conclusion is reinforced by Figure~\ref{fig:neigh_comp}, which shows the box plots of the distributions of $\mathit{fidelity}$, $\mathit{l{\text -}fidelity}$ and $\mathit{cl\text{-}fidelity}$, and some summary data on the size of decision trees ($|c|$), of decision rule premises ($|p|$), and of the number of falsified split conditions in counterfactual rules ($\mathit{nf}$). 
\textbf{LORE} has the highest mean and median f1-measures (high mimic of the black box), the smallest interquartile ranges (low variability of results), and the lowest complexity sizes.
Only for $\mathit{cl\text{-}fidelity}$ \textbf{LORE} has the largest variability, but a median value that is higher than the $90^{th}$ percentile of the competitors. 

\subsection{Comparison with the State-of-Art}
\label{sec:lore_vs_lime}

In this section we compare our approach with the state of the art.  

\subsubsection{Rules vs Linear Regression for Explanations}
We present first a quantitative and qualitative comparison with the linear explanations of LIME\footnote{\url{https://github.com/marcotcr/lime}}~\cite{ribeiro2016should}.
A first crucial difference is that in LIME, the number of features composing an explanation is an input parameter that must be specified by the user.
\textbf{LORE}, instead, automatically provides the user with an explanation including only the features useful to justify the black box decision. 
This is a clear improvement over LIME. 
In experiments, unless otherwise stated, we vary the number of features of LIME explanations from two to ten and we consider the performance with the highest score.

\begin{table}[t]
\setlength{\tabcolsep}{0.5mm}
\centering
\small
\begin{tabular}{c|cc|cc|cc}
\toprule
\textit{Dataset}   & \multicolumn{2}{c|}{german} & \multicolumn{2}{c|}{compass}     & \multicolumn{2}{c}{adult}        \\
\hline
\textit{Black Box} & lore           & lime      & lore           & lime           & lore           & lime      \\
\midrule
\textit{RF}        & \textbf{.925 $\pm$ .2} & .880 $\pm$ .3 & \textbf{.941 $\pm$ .2} & .826 $\pm$ .4 & \textbf{.901 $\pm$ .3} & .824 $\pm$ .4 \\
\textit{NN}        & .980 $\pm$ .1 & \textbf{1.00 $\pm$ .0} & \textbf{.987 $\pm$ .1} & .902 $\pm$ .3 & .918 $\pm$ .3 & \textbf{.998 $\pm$ .1} \\
\textit{SVM}       & \textbf{1.00 $\pm$ .0} & .966 $\pm$ .1 & \textbf{.997 $\pm$ .1} & .900 $\pm$ .3 & .985 $\pm$ .1 & \textbf{.987 $\pm$ .1} \\
\bottomrule
\end{tabular}
\caption{\textbf{LORE} vs LIME: $\mathit{hit}$ scores.}
\label{tab:comp_lime_hit}
\vspace{-5mm}
\end{table}

\begin{figure*}[t]
    \centering
    \hspace{-2mm}
    	\includegraphics[trim = 2mm 0mm 1mm 0mm, clip,width=0.33\linewidth]{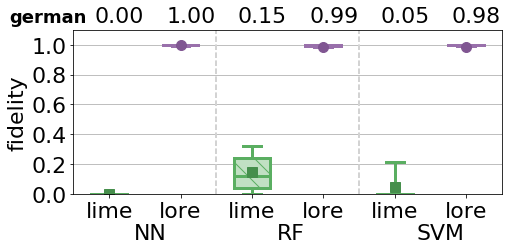} 
   		\includegraphics[trim = 2mm 0mm 1mm 0mm, clip,width=0.33\linewidth]{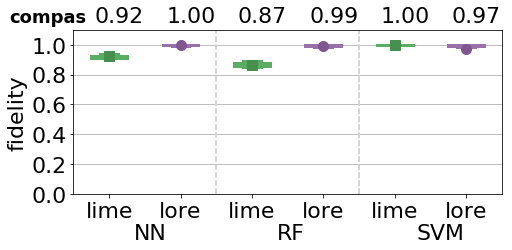}
        \includegraphics[trim = 2mm 0mm 1mm 0mm, clip,width=0.33\linewidth]{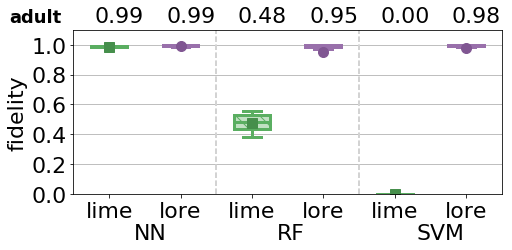}
        \includegraphics[trim = 2mm 0mm 1mm 0mm, clip,width=0.33\linewidth]{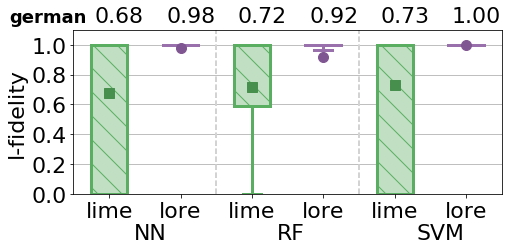} 
   		\includegraphics[trim = 2mm 0mm 1mm 0mm, clip,width=0.33\linewidth]{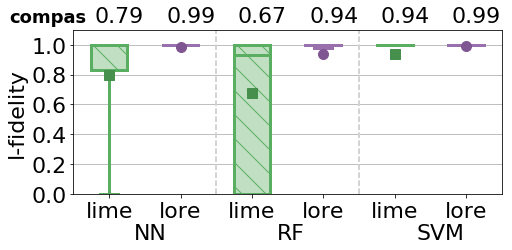}
        \includegraphics[trim = 2mm 0mm 1mm 0mm, clip,width=0.33\linewidth]{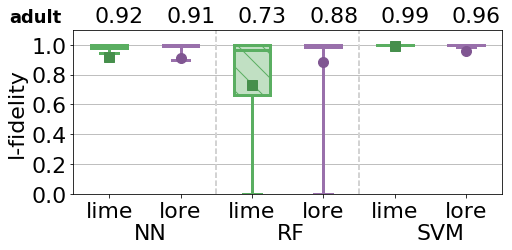}
 	\vspace{-6mm}
\caption{\textbf{LORE} vs LIME: box plots of $\mathit{fidelity}$ and $\mathit{l{\text -}fidelity}$. Numbers on top are the mean values.}
\label{fig:lime_comp}
\vspace{-3mm}
\end{figure*}

\begin{figure}[t]
\centering
\hspace{-2mm}
	\includegraphics[trim = 2mm 0mm 1mm 0mm, clip,width=0.5\linewidth]{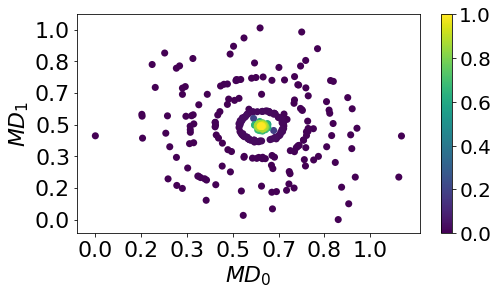} 
   	\includegraphics[trim = 2mm 0mm 1mm 0mm, clip,width=0.5\linewidth]{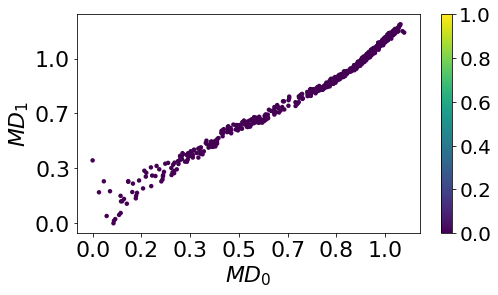}
    \vspace{-6mm}
\caption{Neighborhoods of \textbf{LORE} (left) and LIME (right).}
\label{fig:expl_example_neigh}
\vspace{-2mm}
\end{figure}

\paragraph{Quantitative Comparison}
Table~\ref{tab:comp_lime_hit} reports the mean and standard deviation of $\mathit{hit}$ for each black box predictor and dataset.
Moreover, Figure~\ref{fig:lime_comp} details the box plots of $\mathit{fidelity}$ (top) and $\mathit{l{\text -}fidelity}$ (bottom). 
Results show that \textbf{LORE} definitely outperforms LIME under various viewpoints. 
Regarding the $\mathit{hit}$ score, even when \textbf{LORE} is worse than LIME, it has a score close to $1$. For RF black box, instead, LIME performs considerably worse than \textbf{LORE}. 
The box plots show that, in addition, \textbf{LORE} has better (local) fidelity scores and is more robust than LIME, which, on the contrary, exhibits very high variability in the neighborhood of the instance to explain (i.e.,~for $\mathit{l{\text -}fidelity}$).
%
This can be tracked back to the genetic instance generation of \textbf{LORE}. 
Figure~\ref{fig:expl_example_neigh} reports a multidimensional scaling of the neighborhood of a sample instance $x$ generated by the two approaches. 
\textbf{LORE} computes a dense and compact neighborhood. The instances generated by LIME, instead, can be very distant from each other and always with a low density around $x$. 

\begin{figure}[t]
{\bf - LORE} \hfill
\vspace{1mm}
\begin{minipage}[c]{\linewidth}\em
\begin{tabular}{lp{60ex}}
r = & (\{credit\_amount > 836,
      housing = own, other\_debtors =\\ & \quad none,
      credit\_history = critical account\}
       $\rightarrow$ decision = 0)\\  
$\Phi$ = & \{ (\{credit\_amount $\leq$ 836, 
      housing = own, other\_debtors =\\ & \quad none, 
      credit\_history = critical account\}
     $\rightarrow$ decision = 1),\\
  &   (\{credit\_amount > 836,
       housing = own, other\_debtors =\\ & \quad none, credit\_history = all paid back\}
      $\rightarrow$ decision = 1) \}\\
\end{tabular}
\end{minipage}\\
\vspace{2mm}
{\bf - LIME} \hfill
\vspace{-2mm}
\begin{minipage}[c]{\linewidth}
\centering
\includegraphics[trim = 0mm 0mm 0mm 0mm, clip,width=0.65\linewidth]{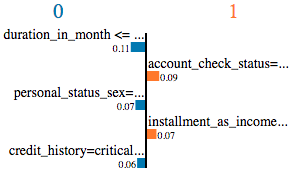}
\end{minipage}\\
\vspace{-2mm}
{\bf - Anchor} \hfill
\vspace{1mm}
\begin{minipage}[c]{\linewidth}\em
\begin{tabular}{lp{60ex}}
a = & (\{credit\_history = critical account,\\ & duration\_in\_month $\in$ [0, 18.00]\}
       $\rightarrow$ decision = 0)
\end{tabular}
\end{minipage}
\caption{Explanations of LORE, LIME and Anchor.}
\label{fig:expl_example}
\vspace{-3mm}
\end{figure}

\paragraph{Qualitative Comparison}
We claim that the explanations provided by \textbf{LORE} are more abstract and comprehensible than the ones of LIME.
Consider the example in Figure~\ref{fig:expl_example}. 
The top part reports a \textbf{LORE} local explanation for an instance $x$ from the \texttt{german} dataset. 
The central part is a LIME explanation. Weights are associated to the categorical values in the instance $x$ to explain, and to continuous upper/lower bounds where the bounding values are taken from $x$. 
Each weight tells the user how much the decision would have changed for different (resp., smaller/greater) values of a specific categorical (resp., continuous) feature. 
In the example, the weight $0.11$ has the following meaning \cite{ribeiro2016should}: if the duration in months had been higher than the value it is for $x$, the prediction would have been, on average, 0.11 less ``0'' (or 0.11 more ``1'').
A not very easy logic to follow when compared to a single decision rule which characterize the contextual conditions for the decision of the black box.
Another major advantage of our notion of explanation consists of the set of counterfactual rules. 
LIME provides a rough indication  of where to look for a different decision: different categorical values or lower/higher continuous values of some feature. 
\textbf{LORE}'s counterfactual rules provide high-level and minimal-change contexts for reversing the outcome prediction of the black box.

\subsubsection{Rules vs Anchors for Explanations} 
A recent extension of LIME is the Anchor\footnote{\url{https://github.com/marcotcr/anchor}} approach \cite{ribeiro2018anchors}. It provides explanations in the form of decision rules, called anchors. Rules are computed by incrementally adding equality conditions in the premise, while an estimate of the rule precision is above a minimum threshold (set to 95\%). Such an estimation relies on neighborhood generation through pure-exploration multi-armed bandit. 

On a qualitative level of comparison, the Anchor approach requires the \textit{apriori} discretization of continuous features, while the decision rule of \textbf{LORE} benefits of the capabilities of decision tree to split continuous features. Contrast, for instance, the example rules in Figure~\ref{fig:expl_example}. Moreover, the approach of Anchor does not clearly extend to compute counterfactuals. 

\begin{figure*}[t]
    \centering
    \hspace{-2mm}
    	\includegraphics[trim = 2mm 0mm 1mm 0mm, clip,width=0.33\linewidth]{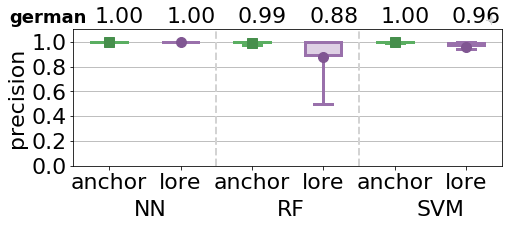} 
   		\includegraphics[trim = 2mm 0mm 1mm 0mm, clip,width=0.33\linewidth]{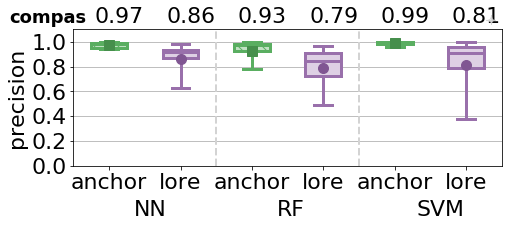}
        \includegraphics[trim = 2mm 0mm 1mm 0mm, clip,width=0.33\linewidth]{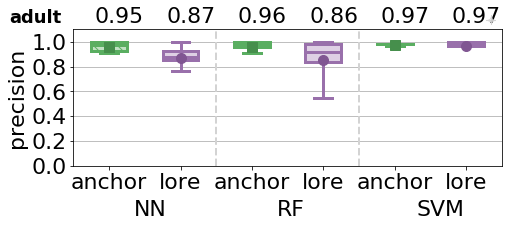}
        \includegraphics[trim = 2mm 0mm 1mm 0mm, clip,width=0.33\linewidth]{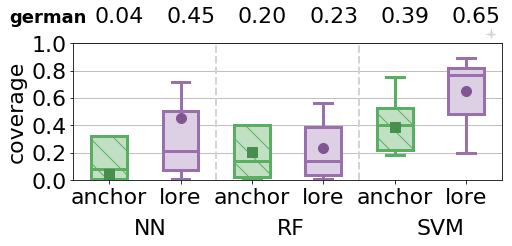} 
   		\includegraphics[trim = 2mm 0mm 1mm 0mm, clip,width=0.33\linewidth]{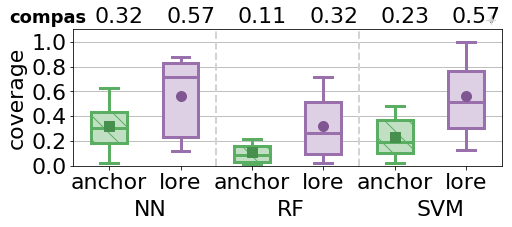}
        \includegraphics[trim = 2mm 0mm 1mm 0mm, clip,width=0.33\linewidth]{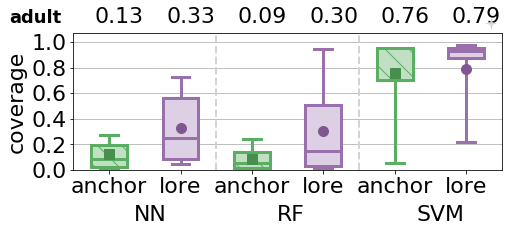}
 	\vspace{-8mm}
\caption{LORE vs Anchor: box plots of \emph{precision} and \emph{coverage}. Numbers on top are the mean values.
\label{fig:precision_coverage}}
\vspace{-3mm}
\end{figure*}

Let us compare now the two approaches on a quantitative level. 
Figure \ref{fig:precision_coverage} reports the average precision of decision rules, where the precision of a rule is the fraction of instances in the neighborhood set that are correctly classified by the rule. 
Although \textbf{LORE} does not require to set the level of precision as parameter, the rule precision is on average high and very similar to that one obtained by Anchor, which is by construction at least 95\%. This can be attributed to the performances of the decision tree induction algorithm, and of the instance generation procedure which produces balanced neighborhoods $Z_-$ and $Z_+$. 
Figure \ref{fig:precision_coverage} also shows the average coverage of decision rules, where the coverage of a rule is the fraction of instances to explain covered by the rule. 
As reported in \cite{ribeiro2018anchors}, large values of coverage are preferable, since this means that the set of decision rules produced over the instances to explain can be condensed/restricted to a subset of it. 
\textbf{LORE} shows a consistently better coverage than Anchor.
Finally, we compare the stability of the two approaches with respect to randomness introduced in the neighboorhood generation. We measure stability using the Jaccard coefficient of feature sets used in the 10 decision rules computed for a same instances in 10 runs of the system. 
Table \ref{tab:stability} reports mean and standard deviation of the Jaccard coefficient. \textbf{LORE} has a better stability than Anchor for all datasets and black boxes.    



\begin{table}[t]
\setlength{\tabcolsep}{0.5mm}
\centering
\small
\begin{tabular}{c|cc|cc|cc}
\toprule
\textit{Dataset}   & \multicolumn{2}{c|}{german} & \multicolumn{2}{c|}{compass}     & \multicolumn{2}{c}{adult}        \\
\hline
\textit{Black box} & lore           & anchor      & lore           & anchor           & lore           & anchor      \\
\midrule
\textit{RF}        & \textbf{.76 $\pm$ .15} & .61 $\pm$ .15 & \textbf{.75 $\pm$ .12} & .73 $\pm$ .14 & \textbf{.70 $\pm$ .15} & .69 $\pm$ .15 \\
\textit{NN}        & \textbf{.69 $\pm$ .18} & .53 $\pm$ .21 & \textbf{.83 $\pm$ .13} & .79 $\pm$ .16 & \textbf{.81 $\pm$ .12} & .65 $\pm$ .16 \\
\textit{SVM}       & \textbf{.82 $\pm$ .16} & .32 $\pm$ .16 & \textbf{.71 $\pm$ .16} & .70 $\pm$ .20 & \textbf{.87 $\pm$ .14} & .67 $\pm$ .13 \\
\bottomrule
\end{tabular}
\caption{LORE vs Anchor: Jaccard measure of stability.}
\label{tab:stability}
\vspace{-8mm}
\end{table}

\section{Conclusion}
\label{sec:conclusion}
We have proposed a local black box agnostic explanation approach based on logic rules. 
\textbf{LORE} builds an interpretable predictor for a given black box and instance to be explained. 
The local interpretable predictor, a decision tree, is trained on a dense set of artificial instances similar to the one to explain generated by a genetic algorithm. 
The decision tree enables the extraction of a local explanation, consisting of a single rule for the decision and a set of counterfactual rules for the reversed decision.
An ample experimental evaluation of the proposed approach has demonstrated the effectiveness of the genetic neighborhood procedure that leads \textbf{LORE} to outperform the proposals in the state of the art.
A number of extensions and additional experiments can be mentioned as future work. 
%
First, \textbf{LORE} now works tabular data. An interesting  future research direction is to make the method suitable for image and text data, for example by applying a pre-processing step for extracting semantic tags/concepts that may be mapped to a tabular format.
Second, another study might be focused on the possibility to derive a global description of the black box bottom-up by composing the local explanations and minimizing the size (complexity) of the global description. 
Third, research lab experiments would be useful for evaluating the human comprehensibility of the provided explanations. 
%
Finally, \textbf{LORE} explanations can be used for identifying data and/or algorithmic biases. 
After the local explanations are retrieved, it would be interesting to develop an approach for deriving an unbiased dataset for safely training the obscure classifier, or to prevent the black box from introducing an algorithmic bias.



\bibliographystyle{abbrv} 
\bibliography{biblio}

\begin{thebibliography}{10}

\bibitem{DBLP:journals/jcst/AggarwalCH10}
C.~C. Aggarwal, C.~Chen, and J.~Han.
\newblock The inverse classification problem.
\newblock {\em J. Comput. Sci. Technol.}, 25(3):458--468, 2010.

\bibitem{andrews1995survey}
R.~Andrews, J.~Diederich, and A.~B. Tickle.
\newblock Survey and critique of techniques for extracting rules from trained
  artificial neural networks.
\newblock {\em Knowl.-Based Syst.}, 8(6):373--389, 1995.

\bibitem{augasta2012reverse}
M.~G. Augasta and T.~Kathirvalavakumar.
\newblock Reverse engineering the neural networks for rule extraction in
  classification problems.
\newblock {\em Neural Processing Letters}, 35(2):131--150, 2012.

\bibitem{back2000evolutionary}
T.~B{\"a}ck, D.~B. Fogel, and Z.~Michalewicz.
\newblock {\em Evolutionary computation 1: Basic algorithms and operators},
  volume~1.
\newblock CRC press, 2000.

\bibitem{baluja1994population}
S.~Baluja.
\newblock Population-based incremental learning. a method for integrating
  genetic search based function optimization and competitive learning.
\newblock Technical report, Carnegie-Mellon Univ Pittsburgh Pa Dept Of Computer
  Science, 1994.

\bibitem{Barocas2016}
S.~Barocas and A.~D. Selbst.
\newblock Big data's disparate impact.
\newblock {\em California Law Review}, 104, 2016.

\bibitem{Berk2017}
R.~Berk, H.~Heidari, S.~Jabbari, M.~Kearns, , and A.~Roth.
\newblock Fairness in criminal justice risk assessments: The state of the art.
\newblock {\em arXiv preprint arXiv:1703.09207}, 2017.

\bibitem{cano2005stratification}
J.~R. Cano, F.~Herrera, and M.~Lozano.
\newblock Stratification for scaling up evolutionary prototype selection.
\newblock {\em Pattern Recognition Letters}, 26(7):953--963, 2005.

\bibitem{craven1996extracting}
M.~Craven and J.~W. Shavlik.
\newblock Extracting tree-structured representations of trained networks.
\newblock In {\em {NIPS}}, pages 24--30. {MIT} Press, 1995.

\bibitem{DBLP:journals/ijamc-igi/DerracGH10}
J.~Derrac, S.~Garc{\'{\i}}a, and F.~Herrera.
\newblock A survey on evolutionary instance selection and generation.
\newblock {\em Int. J. of Applied Metaheuristic Computing}, 1(1):60--92, 2010.

\bibitem{doshi2017towards}
F.~Doshi-Velez and B.~Kim.
\newblock Towards a rigorous science of interpretable machine learning.
\newblock {\em arXiv preprint arXiv:1702.08608v2}, 2017.

\bibitem{eshelman1991chc}
L.~J. Eshelman.
\newblock The chc adaptive search algorithm: How to have safe search when
  engaging in nontraditional genetic recombination.
\newblock In {\em Foundations of genetic algorithms}, volume~1, pages 265--283.
  Elsevier, 1991.

\bibitem{DEAP_JMLR2012}
F.-A. Fortin, F.-M. {De Rainville}, M.-A. Gardner, M.~Parizeau, and C.~Gagn\'e.
\newblock {DEAP}: Evolutionary algorithms made easy.
\newblock {\em Journal of Machine Learning Research}, 13:2171--2175, 2012.

\bibitem{DBLP:journals/kais/FuZL13}
Y.~Fu, X.~Zhu, and B.~Li.
\newblock A survey on instance selection for active learning.
\newblock {\em Knowl. Inf. Syst.}, 35(2):249--283, 2013.

\bibitem{fung2005rule}
G.~Fung, S.~Sandilya, and R.~B. Rao.
\newblock Rule extraction from linear support vector machines.
\newblock In {\em {KDD}}, pages 32--40. {ACM}, 2005.

\bibitem{goodman2016eu}
B.~Goodman and S.~R. Flaxman.
\newblock {EU} regulations on algorithmic decision-making and a "right to
  explanation".
\newblock volume abs/1606.08813, 2016.

\bibitem{guidotti2018survey}
R.~Guidotti, A.~Monreale, F.~Turini, D.~Pedreschi, and F.~Giannotti.
\newblock A survey of methods for explaining black box models.
\newblock {\em arXiv preprint arXiv:1802.01933}, 2018.

\bibitem{hara2016making}
S.~Hara and K.~Hayashi.
\newblock Making tree ensembles interpretable.
\newblock {\em arXiv preprint arXiv:1606.05390}, 2016.

\bibitem{henelius2014peek}
A.~Henelius, K.~Puolam{\"a}ki, H.~Bostr{\"o}m, L.~Asker, and P.~Papapetrou.
\newblock A peek into the black box: exploring classifiers by randomization.
\newblock {\em Data mining and knowledge discovery}, 28(5-6):1503--1529, 2014.

\bibitem{holland1992adaptation}
J.~H. Holland.
\newblock {\em Adaptation in natural and artificial systems: an introductory
  analysis with applications to biology, control, and artificial intelligence}.
\newblock MIT press, 1992.

\bibitem{johansson2004accuracy}
U.~Johansson, L.~Niklasson, and R.~K{\"o}nig.
\newblock Accuracy vs. comprehensibility in data mining models.
\newblock In {\em Int. Conf. on {I}nformation {F}usion}, pages 295--300, vol.
  1, 2004.

\bibitem{krishnan1999extracting}
R.~Krishnan, G.~Sivakumar, and P.~Bhattacharya.
\newblock Extracting decision trees from trained neural networks.
\newblock {\em Pattern recognition}, 32(12), 1999.

\bibitem{lakkaraju2016interpretable}
H.~Lakkaraju, S.~H. Bach, and J.~Leskovec.
\newblock Interpretable decision sets: {A} joint framework for description and
  prediction.
\newblock In {\em {KDD}}, pages 1675--1684. {ACM}, 2016.

\bibitem{lou2012intelligible}
Y.~Lou, R.~Caruana, and J.~Gehrke.
\newblock Intelligible models for classification and regression.
\newblock In {\em {KDD}}, pages 150--158. {ACM}, 2012.

\bibitem{malioutov2017learning}
D.~M. Malioutov, K.~R. Varshney, A.~Emad, and S.~Dash.
\newblock Learning interpretable classification rules with boolean compressed
  sensing.
\newblock In {\em Transparent Data Mining for Big and Small Data}, pages
  95--121. Springer, 2017.

\bibitem{DBLP:journals/prl/McCaneA08}
B.~McCane and M.~Albert.
\newblock Distance functions for categorical and mixed variables.
\newblock {\em Pattern Recognition Letters}, 29(7):986--993, 2008.

\bibitem{nunez2002rule}
H.~N{\'u}{\~n}ez, C.~Angulo, and A.~Catal{\`a}.
\newblock Rule extraction from support vector machines.
\newblock In {\em Esann}, pages 107--112, 2002.

\bibitem{DBLP:journals/air/Olvera-LopezCTK10}
J.~A. Olvera{-}L{\'{o}}pez, J.~A. Carrasco{-}Ochoa, J.~F. {Mart{\'{\i}}nez
  Trinidad}, and J.~Kittler.
\newblock A review of instance selection methods.
\newblock {\em Artif. Intell. Rev.}, 34(2):133--143, 2010.

\bibitem{pedreshi2008discrimination}
D.~Pedreschi, S.~Ruggieri, and F.~Turini.
\newblock Discrimination-aware data mining.
\newblock In {\em {KDD}}, pages 560--568. {ACM}, 2008.

\bibitem{ribeiro2016should}
M.~T. Ribeiro, S.~Singh, and C.~Guestrin.
\newblock "{W}hy should {I} trust you?": Explaining the predictions of any
  classifier.
\newblock In {\em {KDD}}, pages 1135--1144. {ACM}, 2016.

\bibitem{ribeiro2018anchors}
M.~T. Ribeiro, S.~Singh, and C.~Guestrin.
\newblock Anchors: High-precision model-agnostic explanations.
\newblock AAAI, 2018.

\bibitem{ruggieri2004yadt}
S.~Ruggieri.
\newblock Yadt: Yet another decision tree builder.
\newblock In {\em Tools with Artificial Intelligence, ICTAI.}, pages 260--265.
  IEEE, 2004.

\bibitem{singh2016programs}
S.~Singh, M.~T. Ribeiro, and C.~Guestrin.
\newblock Programs as black-box explanations.
\newblock {\em arXiv preprint arXiv:1611.07579}, 2016.

\bibitem{tan2016tree}
H.~F. Tan, G.~Hooker, and M.~T. Wells.
\newblock Tree space prototypes: Another look at making tree ensembles
  interpretable.
\newblock {\em arXiv preprint arXiv:1611.07115}, 2016.

\bibitem{tan2005introduction}
P.-N. Tan, M.~Steinbach, and V.~Kumar.
\newblock Introduction to data mining. 1st, 2005.

\bibitem{DBLP:journals/kbs/TsaiEC13}
C.~Tsai, W.~Eberle, and C.~Chu.
\newblock Genetic algorithms in feature and instance selection.
\newblock {\em Knowl.-Based Syst.}, 39:240--247, 2013.

\bibitem{wachter2017right}
S.~Wachter, B.~Mittelstadt, and L.~Floridi.
\newblock Why a right to explanation of automated decision-making does not
  exist in the general data protection regulation.
\newblock {\em International Data Privacy Law}, 7(2):76--99, 2017.

\bibitem{wachter2017counterfactual}
S.~Wachter, B.~Mittelstadt, and C.~Russell.
\newblock Counterfactual explanations without opening the black box: Automated
  decisions and the {GDPR}.
\newblock {\em Harvard Journal of Law \& Technology, Forthcoming}, 2017.

\bibitem{wang2015falling}
F.~Wang and C.~Rudin.
\newblock Falling rule lists.
\newblock In {\em {AISTATS}}, volume~38 of {\em {JMLR} Workshop and Conference
  Proceedings}. JMLR.org, 2015.

\bibitem{geneticselection2009}
S.~Wu.
\newblock {\em Better Decision Tree from Intelligent Instance Selection}.
\newblock {VDM} Verlag, 2009.

\bibitem{wu2006optimal}
S.~Wu and S.~Olafsson.
\newblock Optimal instance selection for improved decision tree induction.
\newblock In {\em IIE Annual Conference. Proceedings}, page~1. Institute of
  Industrial and Systems Engineers (IISE), 2006.

\bibitem{xu2015show}
K.~Xu et~al.
\newblock Show, attend and tell: Neural image caption generation with visual
  attention.
\newblock In {\em ICML}, pages 2048--2057, 2015.

\bibitem{zhou2016learning}
B.~Zhou, A.~Khosla, {\`{A}}.~Lapedriza, A.~Oliva, and A.~Torralba.
\newblock Learning deep features for discriminative localization.
\newblock In {\em {CVPR}}, pages 2921--2929. {IEEE}, 2016.

\end{thebibliography}

\end{document}